# Adaptive Digital Scan Variable Pixels

Sherin Sugathan
Member, IEEE. Enview
R&D Labs, India. Email:
sherin.s.in@ieee.org

Reshma Scaria
Student Intern
Enview R&D Labs, India.
Email: reshmascaria.mannoor@gmail.com

Alex Pappachen James
Senior Member, IEEE. Nazarbayev
University, Kazakhstan. Email:
apj@ieee.org

*Abstract*—The square and rectangular shape of the pixels in the digital images for sensing and display purposes introduces several inaccuracies in the representation of digital images. The major disadvantage of square pixel shapes is the inability to accurately capture and display the details in the objects having variable orientations to edges, shapes and regions. This effect can be observed by the inaccurate representation of diagonal edges in low resolution square pixel images. This paper explores a less investigated idea of using variable shaped pixels for improving visual quality of image scans without increasing the square pixel resolution. The proposed adaptive filtering technique reports an improvement in image PSNR.

## I. INTRODUCTION

Computer based digital image processing typically assumes the capture of an image, its representation and processing dealing with square shaped picture elements. Matrices are widely used in almost every image processing systems [1]–[4] where each element in the matrix is considered a square shaped pixel. The visual quality of an image tends to degrade as we try to construct a high resolution version of the same image from a low resolution image or a set of low resolution images. This problem is generally solved by super-resolution reconstruction [5] of an image where interpolation techniques are mainly used for reconstruction. It can be observed that the super-resolution reconstruction techniques do treats the nature of resulting pixels or pixel blocks as square shaped.

The rest of the paper is organized as Section II discusses all the related work that explores the idea of using variable pixels. Section III illustrates the proposed method for improving methods based on variable pixels. Section IV reports the results obtained followed by a conclusion in Section IV.

## II. RELATED WORK

The idea proposed in [6], [7], discusses about having an variable pixel layout in sensor hardware. This pixel layout was termed as Penrose Pixel and the algorithms based on this approach proved to perform better than those algorithms which uses a regular square pixel layout. This method of using variable shaped pixel layout was able to generate better visual image quality when compared with the regular pixel layout. A rhombus was used as the primitive shape for creating the Penrose tiles. They used two rhombi structures and each of those structures had five different orientations for creating the pixel layout. Unlike a regular pixel layout, the layout in aperiodic penrose tiling had no translational symmetry which means a particular pixel unit is not repeating exactly [7]. In [8], the penrose pixels were applied in color images.

Unlike the square pixel layout, a spiral architecture was proposed in [9], where every pixel had only six neighboring pixels and the approach reported several advantages over the square pixel layout. In [10], some of the drawbacks in the spiral architecture was solved using a virtual hexagonal structure. In [10], the virtual hexagonal structure was used to uniformly separate an image into sub-images. Another hexagonal lattice method is illustrated in [11], which was inspired from the anatomical consideration of the primate's vision system.

Most of the imaging hardware available today are based on square type pixel layouts. The use of square pixels at the hardware level act as a problem to the method illustrated in [10]. That was the reason for using a virtual hexagonal structure only at the algorithm level. For achieving super-resolution, attempts were also made at the sensor level, but most of those methods suffered from problems like shot noise because the main idea was to reduce the sensor size. When we try to reduce the sensor pixel size (naturally increases the number of pixels per unit area), the amount of light falling at a particular pixel also reduces [12] which is the main reason for causing shot noise.

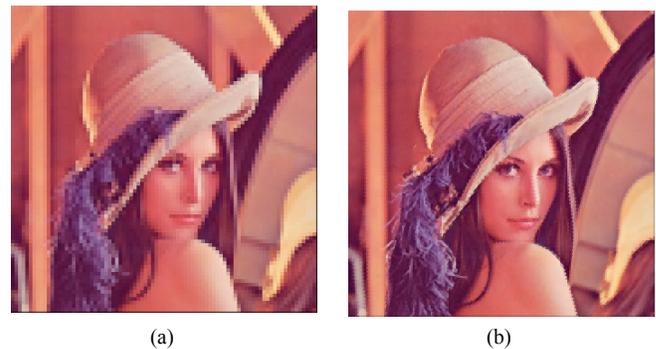

Fig. 1. An image represented using (a) square pixel layout and (b) variable pixel layout.

Inspired from an ancient idea of using mosaic fragments for creating images [13] by the people of Ravenna, Russel [14] proposed the concept of using variable shaped pixels. The main problem Russell cited was the use of square shaped pixels. For improving the viewing accuracy of a digital image, the use of square pixels would demand an increase in the resolution of the image which in turn increases the space requirements along with. However, by using the variable shaped pixel blocks, an image can be represented more accurately than the square type representation.

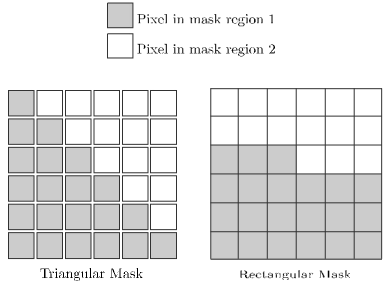

Fig. 2. The two masks (triangular and rectangular) proposed by Russel in [14]. Each of the two masks had two regions indicated as gray and white pixels.

An example showing the difference between the square pixel block representation and an variable pixel block representation is shown in Fig. 1. In [14], Russell proposed the use of two pixel masks, triangular and rectangular as shown in Fig. 2. The size of the masks was chosen as 6 × 6 pixels. The triangular mask consists of two triangles in which one triangle is having 15 pixels and the other is having 21 pixels in it. Similarly for the rectangular mask, there are two rectangle partitions with 15 and 21 pixels. Each of the two masks (triangular and rectangular) were rotated four times by 90 degrees generating a total of eight masks as shown in Fig. 3. Russell replaced all the 6 × 6 pixel block of an image with one of the eight masks for creating an variable pixel image. For every 6 × 6 pixel block in an input image, an appropriate mask is selected by calculating the mean difference between the pixels falling in two regions of the mask.

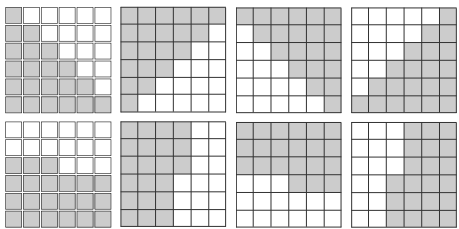

Fig. 3. The eight masks generated by all possible 90 degree rotations of the triangular and rectangular masks.

After computing the difference values for all the eight masks, the 6 × 6 pixel block of the original image is replaced by the mask for which the difference is minimum. This process is repeated for the entire image for generating an variable pixel image. The resulting image can be considered as a combination of triangular and rectangular pixel blocks with different orientations.

In [15], an variable image sampling algorithm is proposed, where the variability is introduced at the sampling stage.

An FPGA based real-time super resolution using an adaptive sensor array is proposed in [16]. They created variable shaped pixels by combining several pixels together as in Russell's method and acknowledges the importance of having variable shaped pixels at the sensor level especially for solving motion blur. The idea proposed by Russell was explored for more types of masks in [17]. In this paper, we aims to bring out few improvements on Russell's idea on variable pixel imaging.

## III. PROPOSED METHOD

The block diagram for the proposed method is shown in Fig. 5. The method shown in Fig. 5 advocates the use of parallel variable blocks to scan the image data. We have used the eight possible orientations (Fig. 3) of the masks inside the scan array. The scan array will generate eight images, where each image corresponds to one of the eight mask orientations. The generated variable images would contain variable blocks generated from the input image data. The formation of an variable block has been depicted in Fig. 4. Each of the parallel scan has a unique distribution of pixel blocks. The pixel blocks used are the same as proposed by Russell in [14]. However, the computation of optimum pixel mask for each image block is not done at the sensor level. This is because it is a difficult and complex approach to have a dynamically changing sensor array, however, it is feasible as a scanning operators on existing pixel arrays.

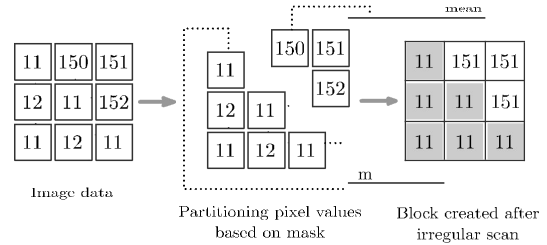

Fig. 4. The formation of an variable pixel block

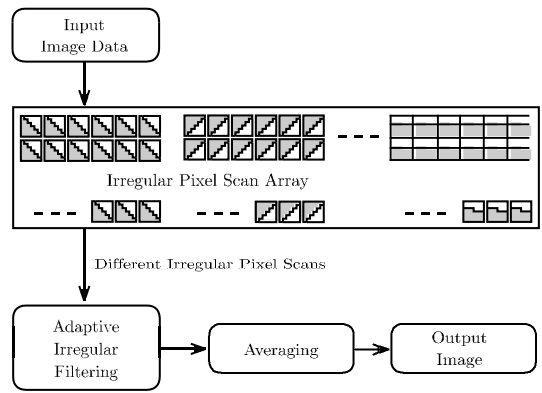

Fig. 5. The proposed variable image scanning system

In Fig. 5, we use only same type of masks in an array instead of computing the optimum mask for every image block. The real image data is captured by eight parallel sensors having the two types of masks, each with four possible rotations.

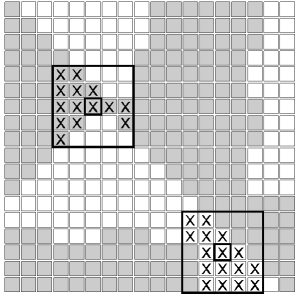

Fig. 6. Two example kernel (5 × 5 pixels) positions over the variable pixel image. The pixels marked as "×" indicates those pixels that together form a new window shape applicable at the center anchor point in that kernel.

Image noise [18]–[20] can be considered as an unwanted signal that affects the quality of an image. In order to address this effect, we have added image noises such as gaussian noise, speckle noise and salt & pepper noise for simulating the impact of sensor noise. We have made improvements in the filtering stage that comes after the addition of noise. The filtering stage introduced in [17] was not considering the variable pixel nature in the image generated from each of the parallel sensors. In [17], they used a square shaped kernel for smoothing an variable pixel image. Here we have used a shape adaptive filter in the filtering stage. Instead of using a regular 5 × 5 pixel smoothing window, the candidate pixels for the smoothing operation is selected in such a way that only the pixels falling in the variable pixel block region where the anchor pixel lies are selected. As you can see in Fig. 6, the pixels marked as "×" are the candidate pixels for computing the smoothed value and they collectively indicate the new shape of the kernel window. For computing the smoothed or filtered value for a particular pixel, only the pixel values marked as "×" will be taken in account for computing the smoothed pixel value. An algorithm for the proposed variable adaptive filtering has been given in Algorithm. 1.

## IV. RESULTS AND DISCUSSION

The proposed parallel variable pixel array system for generating images gives an improved visual quality for the images studied. We have taken a set of 10 standard images (shown in Fig. 7) for the experiments. The images considered are of size 512 × 512 pixels and they are applied to the square shaped and variable shaped parallel sensor arrays. The quality of the generated images is measured by computing the PSNR of the image. The results obtained for various noises like salt & pepper, Gaussian and Speckle noise have been shown in Table. I, II and III. It is obvious that the variable pixel approach has a superiority over the square pixel representation in terms of visual image quality.

In the proposed method, we employed an adaptive variable pixel based filtering for improving image quality. It should be noted that for smaller kernel sizes, the results generated by the adaptive filtering method would stay more similar to that generated by the non-adaptive filtering method. This happens because in a smaller kernel of size, say 3 × 3 pixels, the

## Algorithm 1 Variable Adaptive Filtering

1: procedure VARIABLEFILTER($\mathbf{I}, M, k, \mathbf{f}$)
2:     Read variable input image, $\mathbf{I}$.
3:     Read variable binary mask image, $M$.
4:     Read kernel size, k.
5:     Read the boolean value, $\mathbf{f}$ indicating the type of filtering (mean or median).
6:     $\mathbf{I_p} \leftarrow$ Padded image created from matrix $\mathbf{I}$.
7:     $\mathbf{M_p} \leftarrow$ Padded image created from matrix $M$.
8:     $w \leftarrow$ width of $\mathbf{I_p}$.
9:     $h \leftarrow$ height of $\mathbf{I_p}$.
10:     for $\mathbf{i} \leftarrow 1$ to $w - \lfloor k/2 \rfloor$ in steps of 1 do
11:         for $\mathbf{j} \leftarrow 1$ to $h - \lfloor k/2 \rfloor$ in steps of 1 do
12:             $\mathbf{I_{sub}} \leftarrow k \times k$ neighborhood at $\mathbf{I_p}(\mathbf{i} + (k-1), \mathbf{j} + (k-1))$
13:             $\mathbf{M_{sub}} \leftarrow k \times k$ neighborhood at $\mathbf{M_p}(\mathbf{i} + (k-1), \mathbf{j} + (k-1))$
14:             anchor $\leftarrow \mathbf{M_p}(\mathbf{i} + \lfloor k/2 \rfloor, \mathbf{j} + \lfloor k/2 \rfloor)$
15:             $v \leftarrow \{\mathbf{I_{sub}}(x,y) : (x,y) \in \{(p,q) : \mathbf{M_{sub}}(p,q) = \text{anchor}\}\}$
16:             if ($f ==$ true) then
17:                 $\mathbf{I_{out}}(\mathbf{i}, \mathbf{j}) = \frac{1}{|v|} \sum_{m=1}^{|v|} v_m$
18:             else
19:                 $\mathbf{I_{out}}(\mathbf{i}, \mathbf{j}) = Med(v)$
20:             end if
21:         end for
22:     end for
23:     return $\mathbf{I_{out}}$
24: end procedure

TABLE I
PSNR COMPARISON WHEN SALT & PEPPER NOISE ADDED

| Image Taken | Square Pixel | | Variable | | Variable | |
|---|---|---|---|---|---|---|
| | Mea | Medi | Mea | Medi | Mea | Media |
| Cameram | 23.41 | 23.08 | 24.66 | 25.30 | 25.15 | 25.438 |
| House | 26.69 | 25.73 | 28.37 | 28.68 | 28.51 | 28.805 |
| Jetplane | 22.54 | 21.87 | 23.88 | 24.06 | 24.21 | 24.363 |
| La | 22.09 | 21.62 | 23.15 | 23.54 | 23.54 | 23.721 |
| Le | 24.79 | 24.20 | 25.95 | 26.32 | 26.25 | 26.392 |
| Livingroo | 22.57 | 22.13 | 23.55 | 23.80 | 23.97 | 24.103 |
| Mandril | 19.99 | 19.85 | 20.41 | 20.58 | 20.70 | 20.811 |
| Pepper | 24.74 | 24.00 | 25.99 | 26.31 | 26.22 | 26.336 |
| Pirate | 23.90 | 23.52 | 24.84 | 25.25 | 25.18 | 25.344 |
| Walkbridg | 21.30 | 21.06 | 22.05 | 22.43 | 22.48 | 22.638 |

adaptive method would be eliminating very few non-candidate pixels. In the case of a larger kernel, say 5 × 5 or 7 × 7 pixels, the probability of inclusion of non-candidate pixels is high and thus their removal. We use larger kernel sizes to achieve a higher suppression of noises. In a normal box filtering technique the increase in kernel size would introduce a blurring effect.

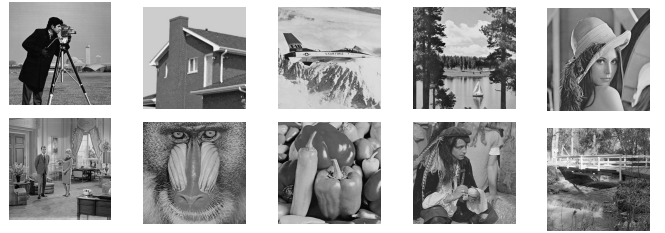

Fig. 7. Set of standard images used for experiments.

If we use a non-adaptive filtering technique, the effect of non-candidate pixels will be high and thus the blurring effect. For an adaptive method, the smoothing is performed in such a manner that the region boundaries or edges represented by variable tiles are preserved. Thus for large kernel sizes, the superiority of adaptive filtering over non-adaptive filtering becomes obvious. The results reported in Table. I, II and III are based on a kernel size of 5 × 5 pixels. The contrast between normal method and the adaptive method would increase for larger kernel sizes. An example output obtained for a square pixel based and variable pixel based image formation system in shown in Fig. 8.

TABLE II
PSNR COMPARISON WHEN GAUSSIAN NOISE ADDED

| Image Taken | Square Pixel | | variable | | variable | |
|---|---|---|---|---|---|---|
| | Mea | Medi | Mea | Medi | Mea | Media |
| Cameram | 23.55 | 23.29 | 24.87 | 24.97 | 25.42 | 25.49 |
| House | 27.14 | 26.17 | 29.05 | 28.26 | 29.26 | 28.92 |
| Jetplane | 22.79 | 22.05 | 24.26 | 23.79 | 24.63 | 24.44 |
| La | 22.30 | 21.99 | 23.49 | 23.38 | 23.92 | 23.82 |
| Le | 25.05 | 24.43 | 26.31 | 25.93 | 26.62 | 26.43 |
| Livingroo | 22.67 | 22.22 | 23.71 | 23.44 | 24.15 | 24.08 |
| Mandril | 20.03 | 19.84 | 20.47 | 20.35 | 20.78 | 20.75 |
| Pepper | 24.97 | 24.32 | 26.34 | 25.96 | 26.60 | 26.48 |
| Pirate | 24.11 | 23.66 | 25.11 | 24.84 | 25.48 | 25.37 |
| Walkbridg | 21.42 | 21.24 | 22.21 | 22.16 | 22.66 | 22.65 |

TABLE III
PSNR COMPARISON WHEN SPECKLE NOISE ADDED

| Image Taken | Square Pixel | | variable | | variable | |
|---|---|---|---|---|---|---|
| | Mea | Medi | Mea | Medi | Mea | Medi |
| Cameram | 23.65 | 23.11 | 25.02 | 24.73 | 25.55 | 25.19 |
| House | 26.88 | 25.83 | 28.64 | 27.56 | 28.83 | 28.28 |
| Jetplane | 22.64 | 21.80 | 24.06 | 23.30 | 24.41 | 23.98 |
| La | 22.26 | 21.85 | 23.41 | 23.16 | 23.85 | 23.60 |
| Le | 24.99 | 24.31 | 26.26 | 25.69 | 26.56 | 26.15 |
| Livingroo | 22.67 | 22.13 | 23.72 | 23.29 | 24.17 | 23.91 |
| Mandril | 20.03 | 19.81 | 20.48 | 20.30 | 20.78 | 20.68 |
| Pepper | 25.04 | 24.16 | 26.45 | 25.72 | 26.69 | 26.19 |
| Pirate | 24.11 | 23.61 | 25.11 | 24.71 | 25.48 | 25.21 |
| Walkbridg | 21.37 | 21.22 | 22.14 | 22.08 | 22.59 | 22.56 |

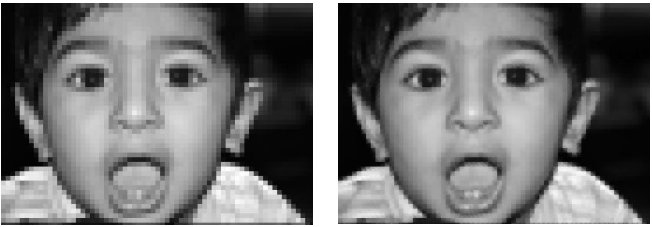

(a) Image generated using square pixel based representation
(b) Image generated for the proposed variable pixel approach

Fig. 8. An example output obtained for (a) square pixel based representation and an (b) variable pixel(adaptive) based representation.

## V. CONCLUSION

In this study, an improved version of an variable pixel based image formation system is presented. It was demonstrated that the filtering techniques can also be adapted according to the variable pixels for improving viewing accuracy. In hardware, variable pixels can be employed to reduce cost, while in software it helps in image compression, which is not studied in this paper. The use of variable pixels for image creation and processing is a promising idea for the future imaging technologies.

ACKNOWLEDGMENT

The discussions with Russell A. Kirsch, the inventor of first digital scan, with National Institute of Technology is acknowledged. The authors would like to thank the support of research grant from Enview Research and Development Labs, Trivandrum, India.